# Making Decisions with Belief Functions

Thomas M. Strat

Artificial Intelligence Center
SRI International
333 Ravenswood Avenue
Menlo Park, California 94025

## Abstract

A primary motivation for reasoning under uncertainty is to derive decisions in the face of inconclusive evidence. However, Shafer's theory of belief functions, which explicitly represents the underconstrained nature of many reasoning problems, lacks a formal procedure for making decisions. Clearly, when sufficient information is not available, no theory can prescribe actions without making additional assumptions. Faced with this situation, some assumption must be made if a clearly superior choice is to emerge. In this paper we offer a probabilistic interpretation of a simple assumption that disambiguates decision problems represented with belief functions. We prove that it yields expected values identical to those obtained by a probabilistic analysis that makes the same assumption. In addition, we show how the decision analysis methodology frequently employed in probabilistic reasoning can be extended for use with belief functions.

## 1 Introduction

Shafer's mathematical theory of evidence [10, 9] (also known as belief functions) has been proposed as the basis for representing and deriving beliefs in computer programs that reason with uncertain information. The ability of the theory to represent degrees of uncertainty as well as degrees of imprecision allows an expert system to represent beliefs more accurately than it could using only a probability distribution. Despite these virtues, the theory of belief functions has lacked a formal basis upon which decisions can be made in the face of imprecision [1]. In contrast, a sizable subfield known as *decision theory* provides a formal basis for decision-making when the underlying representation of uncertainty is ordinary probability [2]. Shafer's "Constructive Decision Theory" addresses the need for judgment at every level of a decision problem, but does not attempt to generalize decision theory for belief functions [11]. More recently, Lesh has proposed a methodology based on an empirically derived coefficient of ignorance whereby clear-cut decisions result [6].

In the present paper we use a theoretically motivated probabilistic assumption to decide among actions when evidence is represented by belief functions. This approach leads to a generalization of decision analysis that is derived from the close relationship between belief functions and probability. The approach offers the foundation for a decision theory for belief functions and provides a formal basis upon which systems that employ belief functions can make decisions.

## 2 Expected Value

Decision analysis provides a methodological approach for making decisions. The crux of the method is that one should choose the action that will maximize the expected value. In this section we review the computation of expected value using a probabilistic representation of a simple example and show how a belief function gives rise to a range of expected values. We then show how a simple assumption about the benevolence of nature leads to a means for choosing a single-point expected value for belief functions.

### 2.1 Expected value using probabilities

**Example – Carnival Wheel #1** A familiar game of chance is the carnival wheel pictured in Figure 1. This wheel is divided into 10 equal sectors, each of which has a dollar amount as shown. For a $6.00 fee, the player gets to spin the wheel and receives the amount shown in the sector that stops at the top. Should we be willing to play?

[1] This research was partially supported by the Defense Advanced Research Projects Agency under Contract No. N00039-88-C-0248 in conjunction with the U.S. Navy Space and Naval Warfare Systems Command.

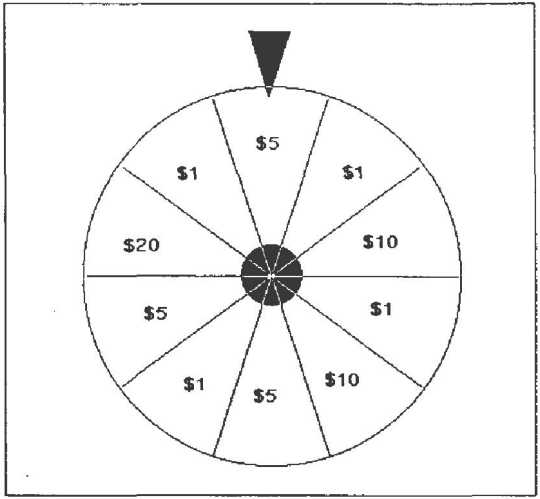

Figure 1: Carnival Wheel #1

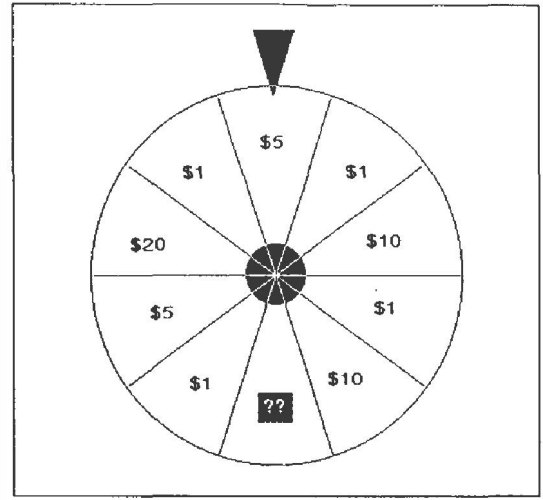

Figure 2: Carnival Wheel #2

The analysis of this problem lends itself readily to a probabilistic representation. From inspection of the wheel (assuming each sector really is equally likely), we can construct the following probability distribution:

$$p(\$1) = 0.4$$
$$p(\$5) = 0.3$$
$$p(\$10) = 0.2$$
$$p(\$20) = 0.1$$

The expected value $E(\Theta)$, where $\Theta$ is the set of possible outcomes, is computed from the formula

$$E(\Theta) = \sum_{a \in \Theta} a \cdot p(a) \quad (1)$$

For the carnival wheel

| $a$ | $p(a)$ | $a \cdot p(a)$ |
|-----|--------|----------------|
| 1   | 0.4    | 0.4            |
| 5   | 0.3    | 1.5            |
| 10  | 0.2    | 2.0            |
| 20  | 0.1    | 2.0            |
| $E(\Theta) =$ | | 5.90 |

Therefore, we should refuse to play, because the expected value of playing the game is less than the $6.00 cost of playing[2]. Let us now modify the problem slightly in order to motivate a belief function approach to the problem.

---

[2]Here we assume that the monetary value is directly proportional to utility because of the small dollar amounts involved.

## 2.2 Expected value using belief functions

**Example – Carnival Wheel #2** Another carnival wheel is divided into 10 equal sectors, each having $1, $5, $10, or $20 printed on it. However, one of the sectors is hidden from view. How much are we willing to pay to play this game?

This problem is ideally suited to an analysis using belief functions. In a belief function representation a unit of belief is distributed over the space of possible outcomes (commonly called the *frame of discernment*). Unlike a probability distribution, which distributes belief over *elements* of the outcome space, this distribution (called a *mass function*) attributes belief to *subsets* of the outcome space. Belief attributed to a subset signifies that there is reason to believe that the outcome will be among the elements of that subset, without committing to any preference among those elements. Formally, a mass distribution $m_\Theta$ is a mapping from subsets of a frame of discernment $\Theta$ into the unit interval:

$$m_\Theta : 2^\Theta \mapsto [0, 1],$$

such that

$$m_\Theta(\phi) = 0 \quad \text{and} \quad \sum_{A \subseteq \Theta} m_\Theta(A) = 1.$$

---

We could instead have chosen to work in a frame of utilities to account for nonlinearities in one's preferences for money. We'll have more to say about utility in the discussion that follows this section.

Any subset that has been attributed nonzero mass is called a *focal element*. One of the ramifications of this representation is that the belief in a hypothesis $A$ ($A \subseteq \Theta$) is constrained to lie within an interval $[Spt(A), Pls(A)]$, where

$$Spt(A) = \sum_{A_i \subseteq A} m_\Theta(A_i) \; ; \; Pls(A) = 1 - Spt(\neg A). \quad (2)$$

These bounds are commonly referred to as *support* and *plausibility*.

The frame of discernment $\Theta$ for Wheel #2 is $\{\$1, \$5, \$10, \$20\}$. The mass function for Wheel #2 is shown below:

$$\begin{aligned} m(\{\$1\}) &= 0.4 \\ m(\{\$5\}) &= 0.2 \\ m(\{\$10\}) &= 0.2 \\ m(\{\$20\}) &= 0.1 \\ m(\{\$1, \$5, \$10, \$20\}) &= 0.1 \end{aligned}$$

and its associated belief intervals are

$$\begin{aligned} [Spt(\{\$1\}), Pls(\{\$1\})] &= [0.4, 0.5] \\ [Spt(\{\$5\}), Pls(\{\$5\})] &= [0.2, 0.3] \\ [Spt(\{\$10\}), Pls(\{\$10\})] &= [0.2, 0.3] \\ [Spt(\{\$20\}), Pls(\{\$20\})] &= [0.1, 0.2] \end{aligned}$$

Before we can compute the expected value of this belief function we must somehow assess the value of the hidden sector. We know that there is a 0.1 chance that the hidden sector will be selected, but what value should we attribute to that sector? If the carnival hawker were allowed to assign a dollar value to that sector, he would surely have assigned $1. On the other hand if we (or a cooperative friend) were allowed to do so, it would have been $20. Any other assignment method would result in a value between $1 and $20, inclusive. Therefore, if we truly do not know what assignment method was used, the strongest statement that we can make is that the value of the hidden sector is between $1 and $20. Using interval arithmetic we can apply the expected value formula of Equation 1 to obtain an *expected value interval* (EVI):

$$E(\Theta) = [E_*(\Theta), E^*(\Theta)] \quad (3)$$

where[3]

$$E_*(\Theta) = \sum_{A \subseteq \Theta} \inf(A) \cdot m_\Theta(A)$$

$$E^*(\Theta) = \sum_{A \subseteq \Theta} \sup(A) \cdot m_\Theta(A)$$

---

[3] We use $\inf(A)$ or $\sup(A)$ to denote the smallest or largest element in the set $A \subseteq \Theta$. $\Theta$ is assumed to be a set of scalar values [13].

The expected value interval of Wheel #2 is

$$E(\Theta) = [5.50, 7.40] \quad (4)$$

As many researchers have pointed out, an interval of expected values is not very satisfactory when we have to make a decision. Sometimes it provides all the information necessary to make a decision, e.g. if the game cost $5 to play then we should clearly be willing to play regardless of who gets to assign a value to the hidden sector. Sometimes we can defer making the decision until we have collected more evidence, e.g. if we could peek at the hidden sector and then decide whether or not to play. But the need to make a decision based on the currently available information is often inescapable, e.g. should we spin Wheel #2 for a $6 fee? We will present our methodology for decision-making using belief functions after pausing to consider a Bayesian analysis of the same situation.

If we are to use the probabilistic definition of expected value from Equation 1, we are forced to assess probabilities of all possible outcomes. To do this, we must make additional assumptions before proceeding further. One possible assumption is that all four values of the hidden sector ($1, $5, $10, $20) are equally likely, and we could evenly distribute among those four values the 0.1 chance that the hidden sector is chosen. This is an example of the generalized insufficient reason principle advanced by Smets [12]. The resulting computation of expected value with this assumption is shown below; its expected value is $6.30:

| $a$ | $p(a)$ | $a \cdot p(a)$ |
|---|---|---|
| 1 | 0.425 | 0.425 |
| 5 | 0.225 | 1.125 |
| 10 | 0.225 | 2.250 |
| 20 | 0.125 | 2.500 |
| $E(\Theta) =$ | | 6.30 |

An alternative assumption is that the best estimate of the probability distribution for the value of the hidden sector is the same as the known distribution of the visible sectors. Using this assumption, the result is $6.00:

| $a$ | $p(a)$ | $a \cdot p(a)$ |
|---|---|---|
| 1 | 4/9 | 4/9 |
| 5 | 2/9 | 10/9 |
| 10 | 2/9 | 20/9 |
| 20 | 1/9 | 20/9 |
| $E(\Theta) =$ | | 6.00 |

Rather than making one of these assumptions, we may wish to parameterize by an unknown probability $p$ our belief that either we get to choose the value of the hidden sector or the carnival hawker does:

**Definition:**
*Let $\rho$ = the probability that the value assigned to the hidden sector is the one that we would have assigned, if given the opportunity.*
*$(1-\rho)$ = the probability that the carnival hawker chose the value of the hidden sector.*

Using this probability we can recompute the expected value of the hidden sector:

$$p(\$20) = \rho$$
$$p(\$1) = 1 - \rho$$
$$E(\text{hidden sector}) = (20)\rho + (1)(1-\rho) = 1 + 19\rho$$

The expected value of Wheel #2 can then be recomputed using probabilities and Equation 1 as illustrated here:

| $a$ | $p(a)$ | $a \cdot p(a)$ |
|---|---|---|
| 1 | $0.4 + 0.1(1-\rho)$ | $0.5 - 0.1\rho$ |
| 5 | 0.2 | 1.0 |
| 10 | 0.2 | 2.0 |
| 20 | $0.1 + 0.1\rho$ | $2.0 + 2\rho$ |
| | $E(\Theta) =$ | $5.50 + 1.90\rho$ |

To decide whether to play the game, we need only assess the probability $\rho$. For the carnival wheel it would be wise to allow that the hawker has hidden the value from our view, thus we might assume that $\rho = 0$. So $E(\Theta) = 5.50$ and we should not be willing to spin the wheel for more than $5.50.

> **Example – Carnival Wheel #3** A third carnival wheel is divided into 10 equal sectors, each having $1, $5, $10, or $20 printed on it. This wheel has 5 sectors hidden from view. However, we do know that none of these sectors is a $20, that the first hidden sector is either a $5 or a $10, and that the second hidden sector is either a $1 or a $10. How much are we willing to pay to spin Wheel #3?

A probabilistic analysis of Wheel #3 requires one to make additional assumptions. Estimating the conditional probability distribution for each hidden sector would provide enough information to compute the expected value of the wheel. Alternatively, estimating just the expected value of each hidden sector would suffice as well. However, this can be both tedious and frustrating: tedious because there may be many hidden sectors, and frustrating because we're being asked to provide information that, in actuality, we do not have. (If we knew the conditional probabilities or the expected values, we would have used them in our original analysis.) What is the minimum information necessary to establish a single expected value for Wheel #3?

The probability, $\rho$, that we used to analyze Wheel #2 can be used here as well. Estimating $\rho$ is sufficient to restrict the expected value of a mass function to a single point. It is easy to see that the expected value derived from this analysis as $\rho$ varies from 0 to 1 is exactly the value obtained by linear interpolation of the EVI that results from using belief functions. The following derivation shows that this is true in general.

**Theorem:** *Given a mass function $m_\Theta$ defined over a scalar frame $\Theta$, and an estimate of $\rho$, the probability that all residual ignorance will turn out favorably, the expected value of $m_\Theta$ is*

$$E(\Theta) = E_*(\Theta) + \rho \cdot (E^*(\Theta) - E_*(\Theta)) \quad (5)$$

**Proof:**
Consider a mass function $m_\Theta$ defined over a frame of discernment $\Theta$. Now consider any focal element $A \subseteq \Theta$, such that $m_\Theta(A) > 0$. Since $\rho$ = the probability that a cooperative agent will control which $a \in A$ will be selected and $(1 - \rho)$ = the probability that an adversary will be in control, then the probability that $a$ will be chosen is

$$p_\Theta(a|A) = \begin{cases} \rho & \text{if } a = \sup(A) \\ (1-\rho) & \text{if } a = \inf(A) \\ 0 & \text{otherwise} \end{cases}$$

Considering all focal elements in $m_\Theta$, we can construct a probability distribution $p_\Theta(a)$ using Bayes' rule:

$$p_\Theta(a) = p_\Theta(a|A) \cdot p_\Theta(A)$$

$$p_\Theta(a) = \sum_{A:\, \sup(A)=a} \rho \cdot m_\Theta(A) + \sum_{A:\, \inf(A)=a} (1-\rho) \cdot m_\Theta(A)$$

Using Equation 1 we have

$$\begin{aligned}
E(\Theta) &= \sum_{a \in \Theta} a \cdot p_\Theta(a) \\
&= \sum_{a \in \Theta} a \cdot \left( \sum_{A:\, \sup(A)=a} \rho \cdot m_\Theta(A) \right. \\
&\qquad \left. + \sum_{A:\, \inf(A)=a} (1-\rho) \cdot m_\Theta(A) \right) \\
&= \sum_{a \in \Theta} \left( \sum_{A:\, \sup(A)=a} \sup(A) \cdot \rho \cdot m_\Theta(A) \right. \\
&\qquad \left. + \sum_{A:\, \inf(A)=a} \inf(A) \cdot (1-\rho) \cdot m_\Theta(A) \right)
\end{aligned}$$

The double summations can be collapsed to a single summation because every $A \subseteq \Theta$ has a unique $\sup(A) \in \Theta$ and a unique $\inf(A) \in \Theta$.

$$\begin{aligned}
E(\Theta) &= \sum_{A \subseteq \Theta} \sup(A) \cdot \rho \cdot m_\Theta(A) + \inf(A) \cdot (1-\rho) \cdot m_\Theta(A) \\
&= \sum_{A \subseteq \Theta} \inf(A) \cdot m_\Theta(A) + \rho \sum_{A \subseteq \Theta} [\sup(A) - \inf(A)] \cdot m_\Theta(A) \\
&= E_*(\Theta) + \rho(E^*(\Theta) - E_*(\Theta))
\end{aligned} \quad (6)$$

□

The important point here is that the Bayesian analysis provides a meaningful way to choose a distinguished point within an EVI. That distinguished point can then be used as the basis of comparison of several choices when their respective EVI's overlap.

### 2.3 Discussion

The value of the result of an action is frequently measured in money (e.g., in dollars), but people often exhibit preferences that are not consistent with maximization of expected monetary value. The theory of *utility* accounts for this behavior by associating for an individual decision-maker a value (measured in *utiles*) with each state $s$, $u = f(s)$, such that maximization of expected utility yields choices consistent with that individual's behavior [3]. Utility theory can satisfactorily account for a person's willingness to expose himself to risk and should be used whenever one's preferences are not linearly related to value. In this paper we do not distinguish between value and utility—the results apply to either metric.

Because of its interval representation of belief, Shafer's theory poses difficulties for a decision-maker who uses it. While a clear choice can always be made when the intervals do not overlap, Lesh [6] has proposed a different method for choosing a distinguished point to use in the ordering of overlapping choices. Lesh makes use of an empirically-derived "ignorance preference coefficient," $\tau$, that is used to compute the distinguished point called "expected evidential belief (EEB)":

$$EEB(A) = \frac{Spt(A) + Pls(A)}{2} + \tau \frac{(Pls(A) - Spt(A))^2}{2}$$

A choice is made by choosing the action that maximizes the "expected evidential value (EEV)"

$$EEV = \sum_{A_i \subseteq \Theta} A_i \cdot EEB(A_i)$$

There are some important differences between Lesh's approach and the present approach for evidential decision-making. The ignorance preference parameter $\tau$ can be seen as a means for interpolating a distinguished value within a *belief* interval $[Spt(A), Pls(A)]$, while the cooperation probability, $\rho$, is used to interpolate within an interval of *expected values* $[E_*(\Theta), E^*(\Theta)]$. Lesh's parameter $\tau$ is empirically derived and has no theoretical underpinning. In contrast the cooperation parameter $\rho$ has been explained as a probability of a comprehensible event—that the residual ignorance will be resolved in one's favor. It leads to a simple procedure involving linear interpolation between bounds of expected value, and is derived from probability theory. The degree to which it matches human decision preferences remains to be determined.

The use of a single parameter to choose a value between two extremes is similar in spirit to the approach taken by Hurwicz with a probabilistic formulation [4]. Rather than computing the expected value of a variable for which a probability distribution is known, Hurwicz suggested that one could interpolate a decision index between two extremes by estimating a single parameter related to the disposition of nature. When this parameter is zero, one obtains the Wald minimax criterion—the assumption that nature will act as strongly as possible against the decision maker [14]. In contrast to the Hurwicz approach in which one ignores the probability distribution and computes a decision index on the basis of the parameter only, in our approach the expected value interval is computed and interpolation between extremes occurs only within the range of residual imprecision allowed by the class of probability distributions represented by a belief function. Thus our approach is identical to the use of expected values when a probability distribution is available; it is identical to Hurwicz's approach when there are no constraints on the distribution; and it combines elements of both when the distribution is known incompletely.

There may be circumstances in which a single parameter is insufficient to capture the underlying structure of a decision problem. In these cases it would be more appropriate to use a different probability to represent the attitude of nature for each source of ignorance. Let $\rho_i$ be the probability that ignorance within $A_i$ will be decided favorably, $\forall A_i, A_i \subseteq \Theta$. Then we obtain

$$\begin{aligned}
E(\Theta) &= \sum_{A_i \subseteq \Theta} \inf(A_i) \cdot m_\Theta(A_i) \\
&+ \sum_{A_i \subseteq \Theta} \rho_i [\sup(A_i) - \inf(A_i)] \cdot m_\Theta(A_i)
\end{aligned}$$

in place of Equation 6.

# 3 Decision Analysis

In the preceding section we have defined the concept of an expected value interval for belief functions and we have shown that it bounds the expected value that would be obtained with any probability distribution consistent with that belief function. Furthermore, we have proposed a parameter (the probability that residual ignorance will be decided in our behalf) that can be used as the basis for computing a unique expected value when the available evidence only warrants bounds on that expected value. In this section we will show how the expected value interval can be used to generalize probabilistic decision analysis.

Decision analysis was first developed by Raiffa [8] as a means by which one could organize and systematize one's thinking when confronted with an important and difficult choice. It's formal basis has made it adaptable as a computational procedure by which computer programs can choose actions when provided with all relevant information. Simply stated, the analysis of a decision problem under uncertainty entails the following steps:

- list the viable options available for gathering information, for experimentation, and for action;
- list the events that may possibly occur;
- arrange the information you may acquire and the choices you may make in chronological order;
- decide the value to you of the consequences that result from the various courses of action open to you; and
- judge what the chances are that any particular uncertain event will occur.

## 3.1 Decision analysis using probabilities

First we will illustrate decision analysis on a problem that can be represented with probabilities to acquaint the reader with the method and terminology.

> **Example – Oil Drilling #1** A wildcatter must decide whether or not to drill for oil. He is uncertain whether the hole will be dry, have a trickle of oil, or be a gusher. Drilling a hole costs $70,000. The payoff for hitting a gusher, a trickle, or a dry hole are $270,000, $120,000, and $0, respectively. At a cost of $10,000 the wildcatter could take seismic soundings that will help determine the underlying geologic structure. The soundings will determine whether the terrain has no structure, open structure, or closed structure.

The experts have provided us with the joint probabilities shown below. We are to determine the optimal *strategy* for experimentation and action.

| State    | No struct | Open | Closed | Marginal |
|----------|-----------|------|--------|----------|
| Dry      | 0.30      | 0.15 | 0.05   | 0.50     |
| Trickle  | 0.09      | 0.12 | 0.09   | 0.30     |
| Gusher   | 0.02      | 0.08 | 0.10   | 0.20     |
| Marginal | 0.41      | 0.35 | 0.24   | 1.00     |

In decision analysis a decision tree is constructed that captures the chronological order of actions and events [5]. A square is used to represent a decision to be made and its branches are labeled with the alternative choices. A circle is used to represent a chance node and its branches are labeled with the conditional probability of each event given that the choices and events along the path leading to the node have occurred.

To compute the best strategy, the tree is evaluated from its leaves toward its root.

- The value of a leaf node is the value (or utility) of the state of nature it represents.
- The value of a chance node is the expected value of the probability distribution represented by its branches as computed using Equation 1.
- The value of a choice node is the maximum of the values of each of its sons. The best choice for the node is denoted by the branch leading to the son with the greatest value. Ties are broken arbitrarily.

This procedure is repeated until the root node has been evaluated. The value of the root node is the expected value of the decision problem; the branches corresponding to the maximal value at each choice node gives the best strategy to follow (i.e. choices to make in each situation).

The evaluated decision tree for the Oil Drilling example is portrayed in Figure 3. It can be seen that the expected value is $22,500 and the best strategy is to take seismic soundings, to drill for oil if the soundings indicate open or closed structure, and not to drill if the soundings indicate no structure.

## 3.2 Decision analysis using belief functions

To use the decision procedure just described, it must be possible to assess the probabilities of all uncertain events. That is, the set of branches emanating from each chance node in the decision tree must represent a probability distribution. In many scenarios, however, estimating these probability distributions is difficult or impossible, and the decision maker is forced to assign probabilities even though he knows they are unreliable. Using belief functions, one need not estimate any probabilities that are not readily available. The representation better reflects the evidence at hand, but the decision analysis procedure cannot be used with the resulting interval representation of belief. In this section we describe a generalization of decision analysis that accommodates belief functions.

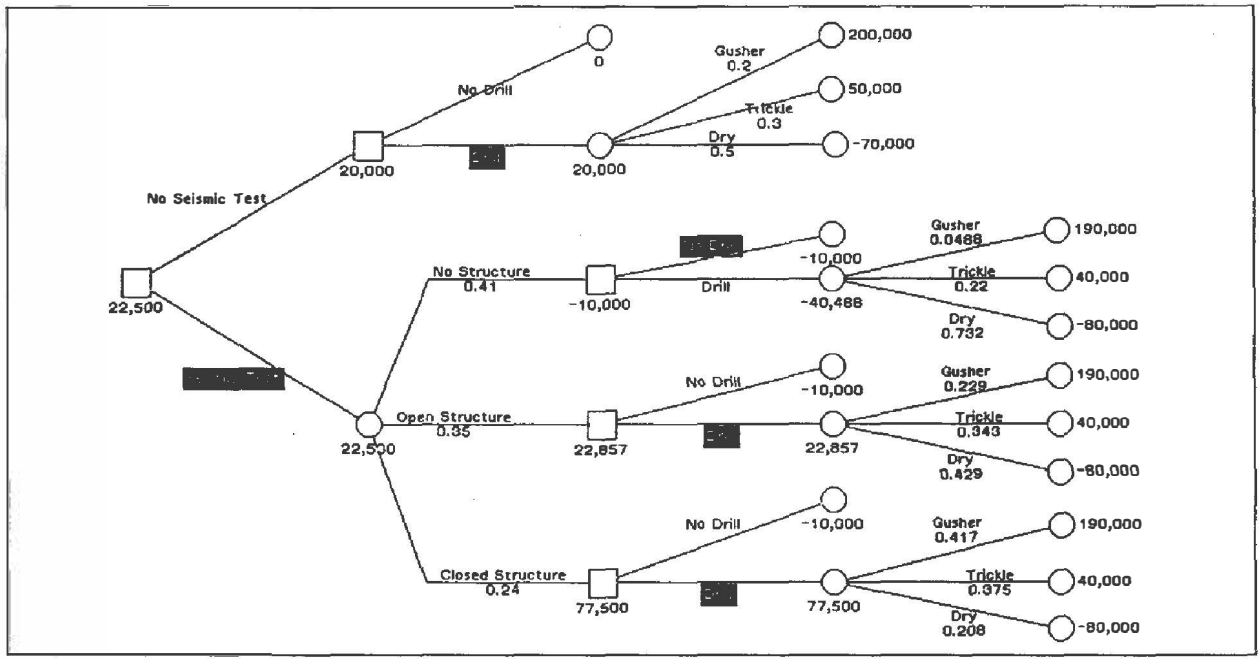

Figure 3: Decision Tree for First Oil Drilling Example

> **Example – Oil Drilling #2** As in the first oil drilling example, a wildcatter must decide whether or not to drill for oil. His costs and payoffs are the same as before: drilling costs $70,000, and the payoff for hitting a gusher, a trickle, or a dry well are $270,000, $120,000, and $0, respectively. However, at this site, no seismic soundings are available. Instead, at a cost of $10,000, the wildcatter can make an electronic test that is related to the well capacity as shown below. We are to determine the optimal strategy for experimentation and action.
>
> | Prob | Test result | Capacity |
> |------|-------------|----------|
> | 0.5  | red         | dry      |
> | 0.2  | yellow      | dry or trickle |
> | 0.3  | green       | trickle or gusher |

Several issues arise that prevent one from constructing a well-formed decision tree for this example. First, consider the branch of the tree in which the test is conducted and the result is green. If we drill for oil, then we know we will find either a trickle or a gusher, but we cannot determine the probability of either from the given information. We are tempted to label the branch with the disjunction, (Trickle ∨ Gusher), with probability 1.0. But, what should be the payoff of that branch? All we can say is that the payoff will be either $40,000 (if a trickle) or $190,000 (if a gusher). Ordinary decision analysis requires a unique value to be assigned, but we have no basis for computing one. So the first modification we make to the construction of decision trees is to allow disjunctions of events on branches emanating from chance nodes, and to allow intervals as the payoffs for leaf nodes. We will discuss later how to evaluate such a tree.

To see the second issue, consider the branch of the tree in which the test is not conducted. If we drill for oil, there is a chance that we will hit a gusher, a trickle, or a dry well, but what is the probability distribution? We know only that

$p(\text{Dry} \mid \text{Red}) = 1.0$  $\qquad p(\text{Red}) = 0.5$
$p(\text{Dry} \vee \text{Trickle} \mid \text{Yellow}) = 1.0$  $\qquad p(\text{Yellow}) = 0.2$
$p(\text{Trickle} \vee \text{Gusher} \mid \text{Green}) = 1.0$  $\qquad p(\text{Green}) = 0.3$

There is not enough information to use Bayes' rule to compute the probability distribution for the well capacity. Without adding a new assumption at this

357

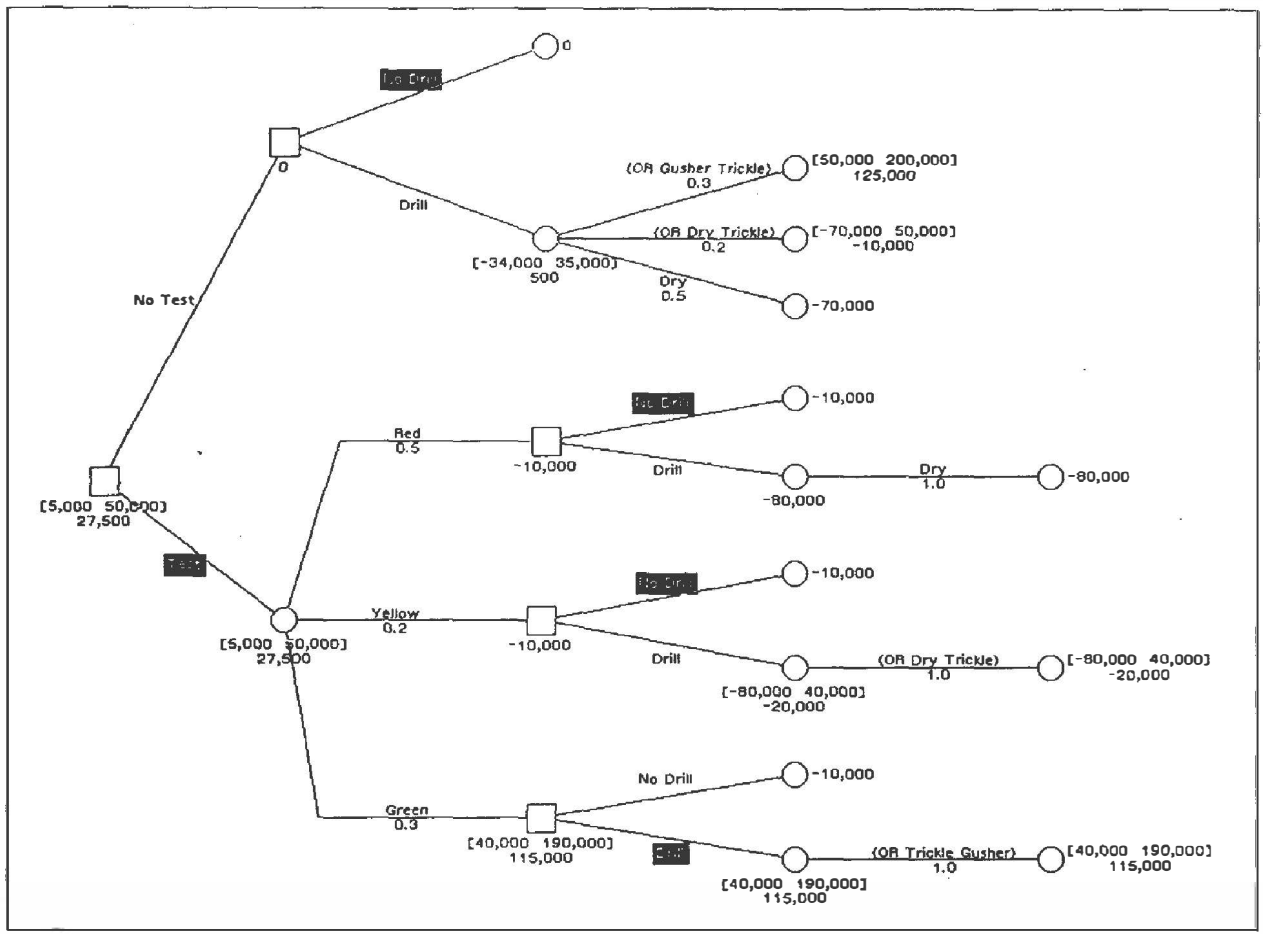

Figure 4: Decision Tree for the Second Oil Drilling Example (assuming $\rho = 0.5$)

point, the strongest statement that can be made is

$$0.5 \leq \quad p(\text{Dry}) \quad \leq 0.7$$
$$0.0 \leq \quad p(\text{Trickle}) \leq 0.5$$
$$0.0 \leq \quad p(\text{Gusher}) \leq 0.3$$

Using belief functions, this can be represented as

$$m(\text{Dry}) = 0.5$$
$$m(\text{Dry} \vee \text{Trickle}) = 0.2$$
$$m(\text{Trickle} \vee \text{Gusher}) = 0.3$$

which yields the required belief intervals

$$[Spt(\text{Dry}), Pls(\text{Dry})] \quad = \quad [0.5, 0.7]$$
$$[Spt(\text{Trickle}), Pls(\text{Trickle})] \quad = \quad [0.0, 0.5]$$
$$[Spt(\text{Gusher}), Pls(\text{Gusher})] \quad = \quad [0.0, 0.3]$$

The second modification we make to decision trees is to allow the branches emanating from a chance node to represent a mass function. The masses must still sum to one, but the events need not be disjoint.[4] The completed decision tree for Oil Drilling Example #2 is shown in Figure 4.

The tools of Section 2 can be used to evaluate a decision tree modified in this manner.

- The value of a leaf node is the value of the state of nature it represents. This may be a unique value or, in the case of a disjunction of states, an interval of values.

- A chance node represents a belief function. Its value is the expected value interval computed with Equation 3

$$E(\Theta) = [E_*(\Theta),\ E^*(\Theta)]$$

- A decision node represents a choice of the several branches emanating from it. The value of each

---
[4]Recall that a probability distribution is an assignment of belief over mutually exclusive elements of a set, whereas a mass function is a distribution over possibly overlapping subsets.

branch may be a point value or an interval. The expected value of an interval is computed using Equation 5 and an estimate of $\rho$

$$E(\Theta) = E_*(\Theta) + \rho \cdot (E^*(\Theta) - E_*(\Theta))$$

The action on the branch that yields the greatest $E(\Theta)$ is chosen. Ties are broken arbitrarily.

Figure 4 shows the evaluated decision tree for $\rho = 0.5$ — each node is labeled with its expected value or expected value interval. In the cases where the expected value is an interval, the evidential expected value $E(\Theta)$ is also shown (using the assumed $\rho$). Preferred decisions are highlighted with a black background.

In summary, a decision tree and decision analysis procedure for belief functions have been described. Two modifications were made to adapt ordinary decision trees: intervals are allowed where values occur; and belief functions are allowed where probability distributions occur. A unique strategy[5] can be obtained by estimating the probability $\rho$.

### 3.3 Comparing two choices

Instead of assuming a $\rho$ value first, and calculating what choices result, one may ask the reverse question. At what value of $\rho$ would I change my decision? This can be answered in general by examining a choice between two states with expected value intervals

Choice 1: $[E_{1*}(\Theta), E_1^*(\Theta)]$
Choice 2: $[E_{2*}(\Theta), E_2^*(\Theta)]$

Using Equation 5 and solving for $\rho$ shows

$$E_1(\Theta) = E_{1*}(\Theta) + \rho \cdot (E_1^*(\Theta) - E_{1*}(\Theta))$$

$$E_2(\Theta) = E_{2*}(\Theta) + \rho \cdot (E_2^*(\Theta) - E_{2*}(\Theta))$$

$$\rho = \frac{E_{1*}(\Theta) - E_{2*}(\Theta)}{(E_2^*(\Theta) - E_1^*(\Theta)) - (E_{2*}(\Theta) - E_{1*}(\Theta))} \quad (7)$$

Letting

$$a = E_{1*}(\Theta) - E_{2*}(\Theta) \text{ and } b = E_2^*(\Theta) - E_1^*(\Theta)$$

gives

$$\rho = \frac{a}{a+b} \quad (8)$$

Thus, Choice 1 is preferable if

$$\rho < \frac{a}{a+b}$$

---
[5] When all values are point-valued and all belief functions are true probability distributions, the strategy will be identical to that prescribed by ordinary decision analysis.

and Choice 2 is preferable if

$$\rho > \frac{a}{a+b}$$

If $\frac{a}{a+b} > 1.0$ then Choice 1 is always preferred (no assumption of $\rho$ is necessary). If $\frac{a}{a+b} < 0.0$ then Choice 2 is always preferred.

One ramification of this decision procedure is that whenever one EVI is slightly "higher" than another, i.e.

$$E_{1*}(\Theta) > E_{2*}(\Theta) \text{ and } E_1^*(\Theta) > E_2^*(\Theta)$$

then it is always preferred. This is because the same value of $\rho$ is assumed to govern the outcome of both choices. Whether this is realistic depends on the situation and deserves further study.

## 4 Discussion

It is interesting to compare the types of assumptions made in a probabilistic analysis with the $\rho$ assumption proposed here for belief functions. When using probability, a maximum entropy assumption is often made. Sometimes, this assumption is justified, and it should properly be considered part of the evidence, not as an assumption. When this is the case, a maximum entropy belief function can be used as well. At other times, the maximum entropy assumption is not justified, but is used simply because *some* assumption must be made, and maximum entropy has some desirable properties [12]. In these cases, the choice of elements in the sample space (the set of possibilities) introduces distortion into the expected value that will result. That is, adding a few more possibilities into the sample space will change the expected value of the maximum entropy distribution over that sample space. (For example, if we chose to allow for the possibility of $2 being among the possibilities for the hidden sector of a carnival wheel the sample space would be $\{1,2,5,10,20\}$ instead of $\{1,5,10,20\}$, and the expected value of the maximum entropy distribution of that sector would be $7.60 instead of $9.00. On the other hand, for any choice of $\rho$ the evidential expected value of the two proceeding sample spaces would be identically $(1 + 19\rho)$. However, adding possibilities outside the interval $[1, 20]$ would change the evidential expected value. For example, allowing for the possibility of $50 in the hidden sector would change the evidential expected value to $(1 + 49\rho)$. The point is that both assumptions introduce bias into the decision criteria. This should not be surprising because both are *unjustified* assumptions after all. There is no basis on which to prefer one over the other; both are entirely plausible.

Having made this point, there are some consequently weak arguments for recommending the use of the assumption of the probability of nature's cooperation $\rho$. Because the EVI spans the range of all expected values that could be obtained by adding *any* assumption to a probabilistic analysis, there always exists some value of $\rho, 0 \leq \rho \leq 1$ that yields the same expected value $E(\Theta)$ as a probabilistic analysis. Therefore, the decisions that are prescribed depend only on one's ability to estimate $\rho$, not on his election to use $\rho$ and Equation 3. Furthermore, the use of a single parameter means that the decision-maker is asked to provide only one additional piece of information.

The parameter $\rho$ has been explained as a probability, giving it a formal grounding that earlier schemes for belief functions have lacked. Furthermore, we feel that it is the probability of a meaningful event. Selecting $\rho = 0$ is appropriate when an adversary controls the situation (as in game playing, for example) or when a decision-maker wishes only to minimize his expected loss. An optimistic decision-maker would prefer to choose $\rho = 1$ to maximize his chance of realizing the greatest possible expected payoff without worrying about what losses he might expect. Intermediate values of $\rho$ can be used to compromise between these extremes.

## 5 Summary

We have described a decision theory for Shafer's theory of belief functions. We started by defining the notion of expected value interval (EVI) and showed it to properly bound the expected value of any probability distribution that could be obtained by introducing additional assumptions. Because an expected value interval is often insufficient for decision-making, we recognize that a point must be chosen to compare alternative choices. We then showed how a linear interpolation of a distinguished value within the EVI is equivalent to making an assumption of the benevolence or maleficence of nature. Letting $\rho$ be the probability that imprecision will be resolved favorably, we derived that distinguished point.

We have also shown how the theory can be used to generalize the decision trees used in probabilistic decision analysis. These tools allow a decision-maker to defer unwarranted assumptions until the latest possible moment. In so doing he can sometimes avoid making any assumptions at all. Otherwise, he is forced to provide only enough additional information to allow a clear choice, and has the benefit of all available information to selectively decide where he would like to make that assumption.